# Semi-supervised learning of images with strong rotational disorder: assembling nanoparticle libraries

Maxim A. Ziatdinov[1,2,a] Muammer Yusuf Yaman,[3] Yongtao Liu,[1,] David Ginger,[3,4] and Sergei V. Kalinin[1]

[1] Center for Nanophase Materials Sciences, Oak Ridge National Laboratory, Oak Ridge, TN 37831

[2] Computational Sciences and Engineering Division, Oak Ridge National Laboratory, Oak Ridge, TN 37831

[3] Department of Chemistry, University of Washington, Seattle, WA, 98195

[4] Physical Sciences Division, Pacific Northwest National Laboratory, Richland, WA, 99354

The proliferation of optical, electron, and scanning probe microscopies gives rise to large volumes of imaging data of objects as diversified as cells, bacteria, pollen, to nanoparticles and atoms and molecules. In most cases, the experimental data streams contain images having arbitrary rotations and translations within the image. At the same time, for many cases, small amounts of labeled data are available in the form of prior published results, image collections, and catalogs, or even theoretical models. Here we develop an approach that allows generalizing from a small subset of labeled data with a weak orientational disorder to a large unlabeled dataset with a much stronger orientational (and positional) disorder, *i.e.*, it performs a classification of image data given a small number of examples even in the presence of a distribution shift between the labeled and unlabeled parts. This approach is based on the semi-supervised rotationally invariant variational autoencoder (ss-rVAE) model consisting of the encoder-decoder "block" that learns a rotationally (and translationally) invariant continuous latent representation of data and a classifier that encodes data into a finite number of discrete classes. The classifier part of the trained ss-rVAE inherits the rotational (and translational) invariances and can be deployed independently of the other parts of the model. The performance of the ss-rVAE is illustrated using the synthetic data sets with known factors of variation. We further demonstrate its application for experimental data sets of nanoparticles, creating nanoparticle libraries and disentangling the representations defining the physical factors of variation in the data. The code reproducing the results is available at https://github.com/ziatdinovmax/Semi-Supervised-VAE-nanoparticles.

[a] ziatdinovma@ornl.gov

The proliferation of optical, electron, and scanning probe microscopies continues to give rise to large volumes of imaging data.[1-4] For biological systems, this includes objects as diversified as cells, bacteria, organelles, and pollen.[5-7] In nanoscience this includes nanoparticles, nanowires, and other nano-objects.[8-10] In electron and scanning probe microscopy, this includes atoms and molecules as detected via changes in the local density of states or nuclei density.[11-14] Correspondingly, there is considerable interest in deriving physical information and potentially actionable insights from these images, ranging from the identification and classification of the defects,[15-17] reconstruction of the statistical and generative[18,19] physical models, or identification of medical conditions and potential interventions.[20] However, in many cases, the data streams coming from the imaging systems contain the objects with a strong translational and rotational disorders, since the objects of interest typically have arbitrary orientation and position in the field of view. For example, in crystalline materials, the preferred orientation is determined by the relevant alignment of the crystallographic axes with respect to the image plane, with potentially small disorder due to local and global strains and scan distortions. At the same time, for mesoscale imaging and disordered materials, the objects of interest can have arbitrary orientation in the image plane. Analysis of these data streams necessitates rapid classification and identification of the observed objects. An often-encountered scenario is the one in which the individual objects are separable, corresponding to the strong dilution of original solution, rare defects, or easily identifiable borders of the objects. In these cases, the compound images containing multiple objects can be separated into the patches containing individual objects of interest, albeit at arbitrary orientation, with positional jitter relative to the center of the patch due to the variability of object shapes. Correspondingly, analysis of such data via supervised or unsupervised machine learning methods needs to account for these factors of variability.

Here, we develop an approach for the special case of such analysis, in which a small number of labels is provided. This corresponds to the practically encountered scenarios where prior data in the form of small manually labeled sets, published papers, catalogs, or other forms is available. The central challenge of such analysis is the generalization from a small subset of labeled data with a weak orientational disorder (e.g., manually labeled data or reference data) to a large unlabeled data set with a much stronger orientational disorder. In the language of deep/machine learning (DL/ML), our goal is to generalize to a dataset characterized by distribution shift, which is one of the key challenges for practical applications of DL/ML models.[21]

To address this challenge, we present an approach for semi-supervised learning based on variational autoencoders with rotational (and translational) invariance. Variational autoencoders (VAE) belong to a class of probabilistic graphical models that can learn the main continuous factors of variation from high-dimensional datasets.[22,23] It consists of generative and inference models of the forms $p(x|z)p(z)$ and $q(z|x)$ usually referred to as decoder and encoder. The former treats the latent variable $z$ as a code from which it tries to reconstruct the observation $x$ using a prescribed prior distribution $p(z)$ (here chosen as standard Gaussian), whereas the latter is used to approximate the usually intractable posterior distribution. Both encoder and decoder are parametrized by deep neural networks. The low-dimensional latent representations learned by VAE can be efficiently used for downstream tasks such as classification even with a limited number of labels. More specifically, to use VAE in a semi-supervised learning setting, one treats a class label as one of the latent factors of variation. This latent variable comes from a discrete distribution and is known *a priori* only for a small fraction of the data. The encoding into the discrete classes for the unlabeled part of data is performed by a separate neural network (*y*-Encoder in Fig. 1), which also plays a role of a classifier. The semi-supervised VAE (ss-VAE) has been successfully applied to standard benchmark datasets such as MNIST[24] where it was able to achieve ~90% accuracy on the dataset with only ~3% of labels available.

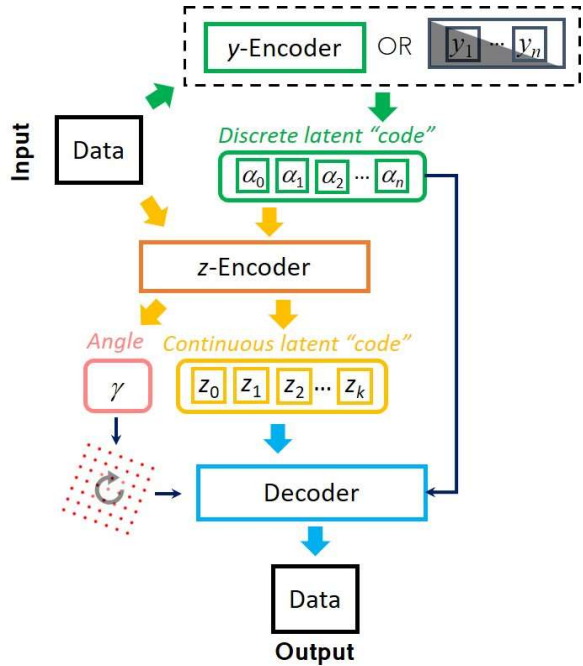

**Figure 1.** Schematic of ss-rVAE model. Here the partially observed labels are treated as a part of the latent "code" that generates the data. For the unlabeled part of the data, the discrete class is encoded using the *y*-Encoder, which is an MLP with the *softmax* activation function at the end. The other latent variables (representing everything but the label) are separated into two parts: the one associated with rotations and translations (for simplicity, only the rotation part is shown) and the one corresponding to the remaining continuous factors of variation. They are encoded using a standard VAE encoder module denoted here as *z*-Encoder. The angle latent variable is used to rotate a pixel grid, which is then concatenated with discrete and continuous latent variables and passed to the VAE's decoder to reconstruct the input data.

Unfortunately, the classical ss-VAE struggles to perform accurate classification of the unlabeled data obtained under conditions different from those of labeled data. For example, as we show below, even a relatively small increase in the orientational disorder for the unlabeled portion of data almost completely throws off the classifier. Simply increasing the fraction of labeled data does not lead to improved results since the "new" labeled data still comes from the same distribution.

To overcome this limitation, we partition a continuous latent space of the ss-VAE into the part associated with orientational and positional disorders and the part corresponding to the remaining factors of variation (shear, scale, etc.). The latent variables from the former are used to rotate (and translate) the pixel grid associated with input images, which is then concatenated with the rest (continuous and discrete) latent variables and passed to the VAE's decoder to enforce a geometric consistency[25] between rotated (and translated) unlabeled objects. As far as the choice of the decoder's prior for the angle latent variable is concerned, we tested both standard normal distribution and intuitively more suitable projected normal distribution[26] (continuous distribution on the circle) and found no significant difference for the datasets used in this paper. The weights of the encoders and decoder are trained by optimizing the standard ss-VAE loss objective with an explicit classification loss for the labeled data.[27]

As a model synthetic data set, we have chosen the dataset with playing card suites as originally introduced in Ref [28]. This choice is predicated on a relatively small number of classes (4), and interesting full and partial degeneracies with respect to the affine transforms. For example, upon compression and 90 degrees rotation the diamond will transform into a (smaller) diamond of the same orientation. For clubs, the rotation by 120 yields an almost identical object shape, allowing to trace the tendency of a model to get captured in metastable minima. Finally, spades and (rotated) hearts differ by the presence of a small tail only.

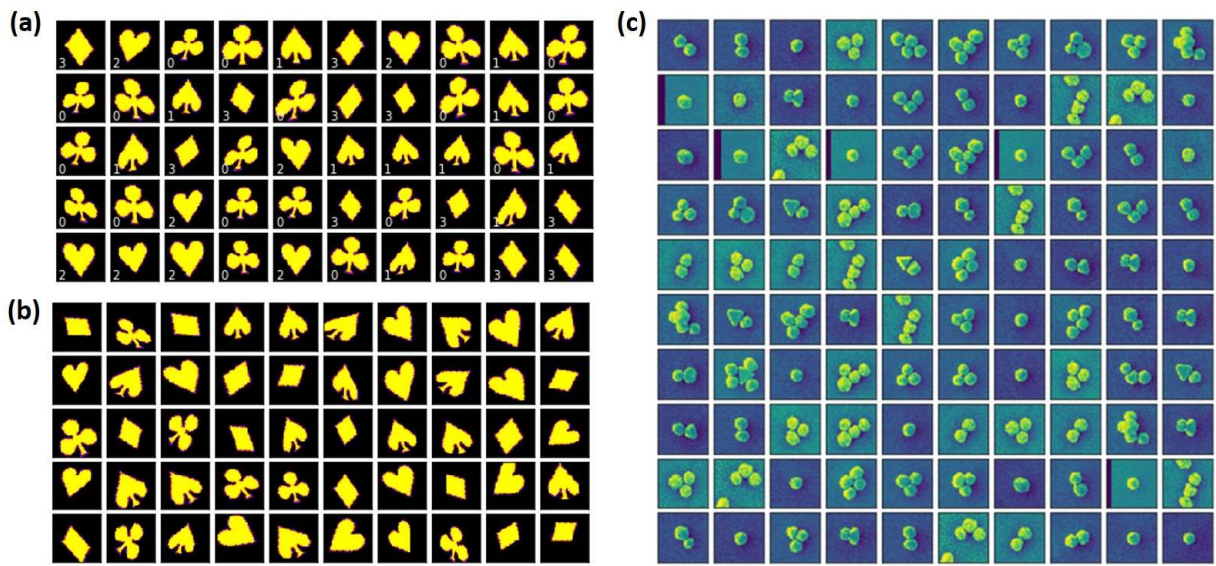

**Figure 2. Illustration of the datasets used in this study.** (a, b) The representation of the cards dataset as introduced in Ref [28]. (a) Typical objects from the labeled part of the dataset with a weak orientational disorder (the rotational angle is sampled uniformly from the [-5°, 5°] range). (b) Typical objects from the unlabeled part of the dataset with a strong orientational disorder (the rotational angle is sampled uniformly from the [-60°, 60°] range. (c) Experimental dataset of the gold nanoparticles.

We generated a data set of $M = 12000$ cards (3000 per each card suite) having the varying angular, translation, shear, and size disorders. In addition, we generated a small number, $N = 800$ (200 per each card suite), of the labeled examples. Importantly, the labeled examples do not have a positional disorder (the objects are fixed at the center of image) and are characterized by only a negligible orientational disorder (the objects are rotated in the range between -5° and 5°) compared to the unlabeled counterpart (see Figure 1a, b). The ss-rVAE training aims to simultaneously address three targets: i) to reconstruct the data set, ii) to establish the structure in the latent space, and iii) to assign the labels to the individual elements of the data set characterized by a distribution shift. Finally, the evaluation of the trained model accuracy was performed on a separate dataset generated using the same disorder parameters as the unlabeled part of the training data but with a different pseudo-random seed. In all the models discussed in the paper, the encoders and decoder were the 2-layer perceptrons with 128 neurons per layer activated by a hyperbolic tangent function, *tanh()*. The weights of the encoders and decoder were trained jointly using the Adam[29] optimizer.

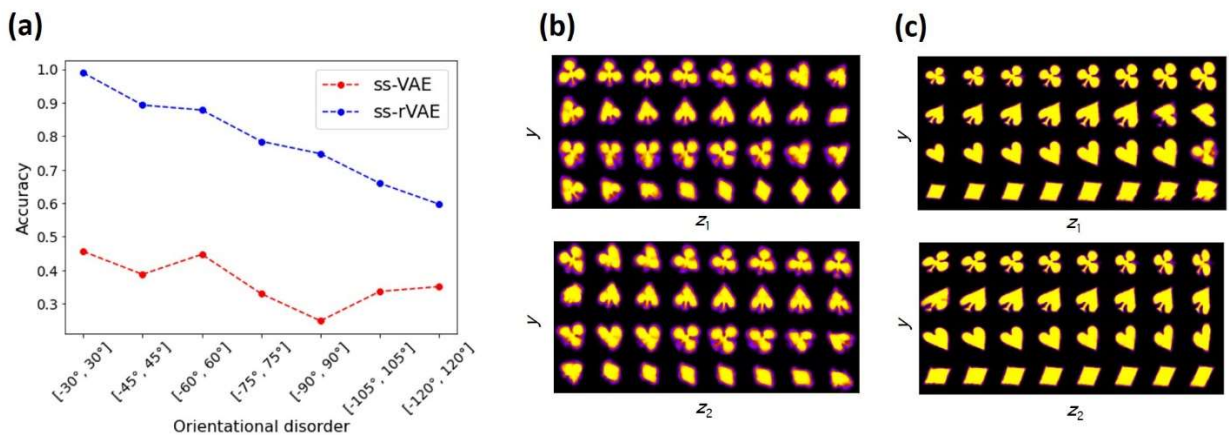

**Figure 3. Performance of the ss-VAE (without rotational invariance) and ss-rVAE (with rotational invariance) on the cards dataset.** (a) Comparison of the classification performance between the regular (ss-VAE) and rotationally invariant (ss-rVAE) models on the unlabeled test datasets characterized by a distribution shift in the form of increasing orientational disorder. The horizontal axis labels show an interval from which the rotation angles for the unlabeled data were sampled uniformly. The labeled part used in training remained the same in all the cases and corresponded to the data described in Figure 2a. (b, c) Latent space traversals of the ss-VAE (b) and ss-rVAE (c) learned from the unlabeled data with rotation angles sampled from [-60°, 60°] interval. Note that the first continuous latent variable (top row in (c)) captures variation in scale whereas the second continuous latent variable (bottom row in (c)) corresponds to a shear strain.

The classification accuracies of the trained ss-VAE (without rotational invariance) and ss-rVAE (with rotational invariance) are shown in Figure 3a for different orientational disorders in the unlabeled dataset. Here, the labels on horizontal axis show a range from which the angles for the unlabeled data were uniformly sampled. For the sake of brevity, we will refer to each distribution by the value of $\alpha$ that forms the [-$\alpha$, $\alpha$] interval. The angles for the labeled data were sampled from a uniform distribution with $\alpha = 5°$ for all the cases. The classification is performed with a y-Encoder (see Fig. 1). Clearly, the ss-VAE fails to generalize to the unlabeled data with different orientations of the same objects, with most of the prediction accuracies just slightly above a random guess. Nor it could learn any meaningful latent representation of the data (Fig. 3b). On the other hand, the ss-rVAE applied to the exact same datasets shows a robust classification performance (> 75% accuracy) for a relatively broad range of the orientational disorder (5° < $\alpha$ < 90°). Furthermore, the ss-rVAE is capable of learning the correct factors of variation for this range of angular disorder (Fig. 3c). Indeed, the first continuous latent variable ($z_1$) clearly captures a variation in scale whereas the second one ($z_2$) encoded a variation in shear deformation. We note that the accompanying Jupyter notebook allows readers to explore different ranges of the disorder parameters as well as to tune the architectures (e.g., change a number of layers in the encoder and decoder modules) of the VAE models.

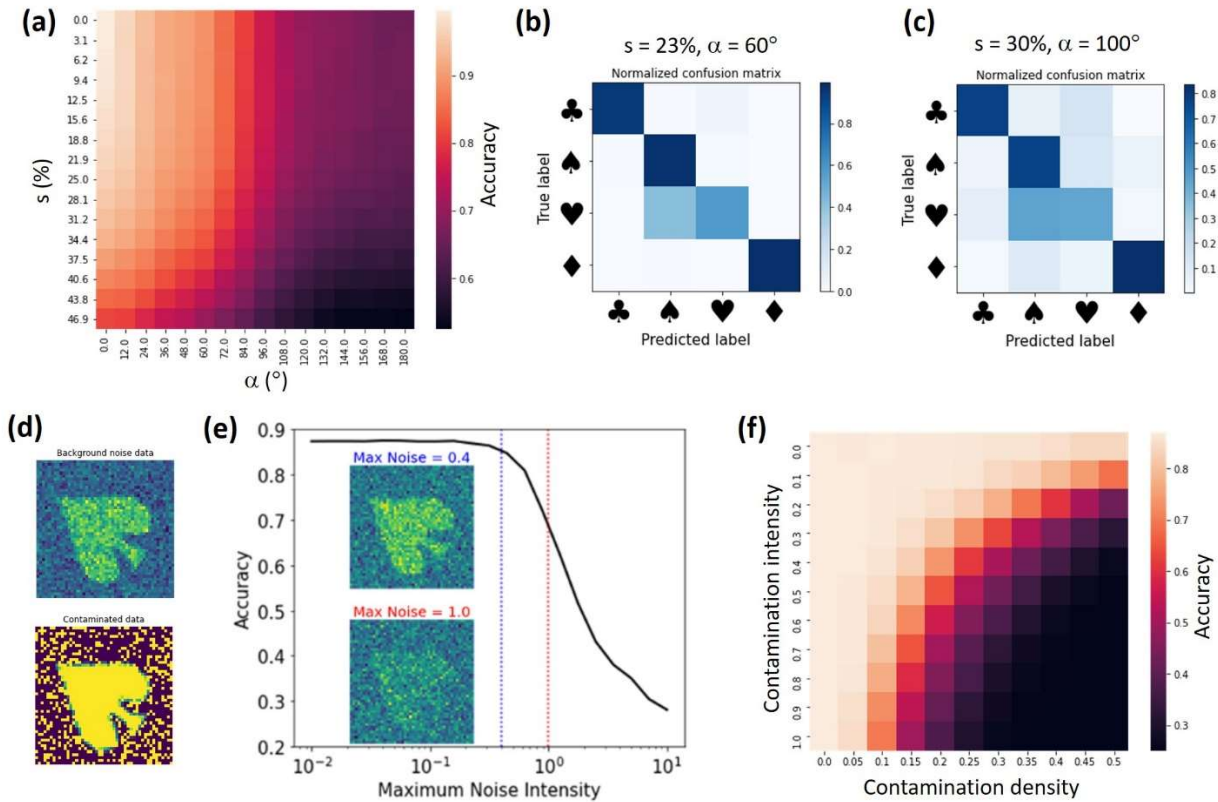

**Figure 4. Performance of the trained ss-rVAE on cards dataset.** (a) The dependence of prediction accuracy on shear ($s$) and angular ($\alpha$) disorders. Here, $s$ denotes the [0, $s$] range from which values of shear deformation are sampled for image transformation. Similarly, $\alpha$ denotes the [-$\alpha$, $\alpha$] range from which the rotation angles are sampled. The model is trained on labeled dataset with $s = 23\%$ and $\alpha = 5°$, whereas the unlabeled part for model training had $s = 23\%$, $\alpha = 60°$,

and random offsets between 0 and 0.1 (in fractions of an image). (b, c) Confusion matrix for test dataset with (b) $s = 23\%$, $\alpha = 60°$, and (c) $s = 30\%$, $\alpha = 100°$. The random offsets are the same as during the training. (d) Examples of data contaminated with two types of noise. (e) The dependence of accuracy on background noise intensity for the test dataset with $s = 23\%$, $\alpha = 60°$. (f) The dependence of accuracy on contamination level for the test dataset with $s = 30\%$, $\alpha = 100°$.

Next, we investigated the performance of the *trained* ss-rVAE model on datasets with varying disorder and noise levels. We note that such "stress tests" are critical for deploying the ML models online (*i.e.*, during the actual experiments) where one may not have time to retrain a model every time there are changes in the data generation process. Figure 4a shows a dependence of ss-rVAE prediction accuracy on angular and shear disorder. Here, we use *s* to denote the [0, *s*] range from which values of shear deformation are uniformly sampled. The ss-rVAE was trained on the dataset with s = 23% and $\alpha = 60°$. One can see that ss-rVAE shows a remarkable robustness even when the shear deformation strength exceeds the maximum shear level used in the training dataset. For the orientational disorder, the accuracy is expectedly decreasing for $\alpha > 60°$, consistent with observations in Fig. 3. Nevertheless, it remains at acceptable levels (> 70%) even at the very high rotation angles (up to $\alpha = 180°$) for the low and moderate shear values. This is remarkable because even though we didn't explicitly enforce the rotational invariance on our classifier (*y*-Encoder), it was learned by the "proximity" to the rVAE part (*z*-Encoder and Decoder) of the model. Finally, for the large shear (> 30%) and strong orientational disorder ($\alpha > 100°$) the model struggles to provide accurate predictions. The confusion matrix analysis of the test datasets (Figure 4b,c) revealed that the misclassification originates mostly from mislabeling hearts to spades, whereas the assignments of all other cards (club, spade, and diamond) remain robust.

To explore noise effects, we introduced two kinds of noise to simulate practical experimental results as demonstrated in Figure 4d. The first one is a simple Gaussian noise. This type of noise is common in experimental measurements and is sometimes referred to as background noise. Figure 4e shows the dependence of the ss-rVAE prediction accuracy on the intensity of the background noise dataset. We can see that ss-rVAE predictions are robust to noise with intensities below 0.4. Moreover, even if the noise intensity reaches 1.0, the ss-rVAE still has an accuracy above *70%* (the insets in 3e show examples of how the data looks like for noise intensities of 0.4 and 1.0). Noteworthily, we believe that most practical experimental data can maintain a noise intensity below 1.0 or even below 0.4, suggesting that ss-rVAE can have a robust performance in practical use. In addition, we note that this ss-rVAE model is trained on a clean dataset, and therefore its performance on noisy data can be further improved if the model is trained using data augmented by noise.

For the second type of noisy data, we added an extra signal intensity to random pixels of the clean data. This type of noise is analogous to measuring experimental samples with contamination, so we refer to this data as contaminated data. We defined two parameters to control the contamination level: one is contamination density that is determined by the ratio of pixels with contamination signal; the other is contamination intensity that is determined by the intensity of the added signal. Shown in Figure 4f is the dependence of ss-rVAE analysis accuracy on contamination level and density. We can see that the ss-rVAE shows good accuracy when the contamination density is below 10%.

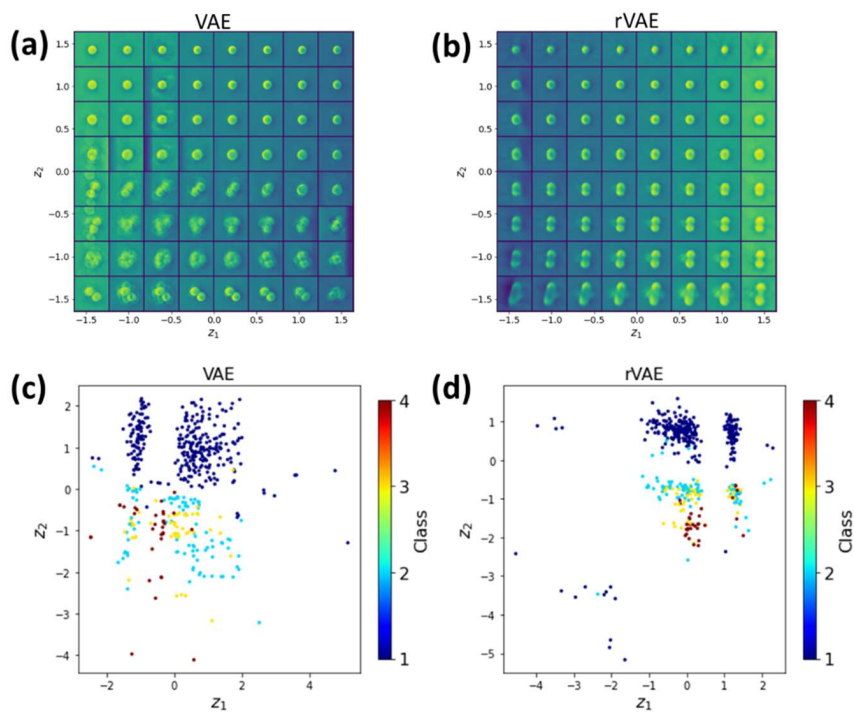

**Figure 5. VAE and rVAE analyses of the nanoparticle dataset. (a).** VAE latent space. **(b)** rVAE latent space. **(c)** VAE latent variables distribution with color corresponding to class. **(d)** VAE latent variables distribution with color corresponding to class. Note that in (c) and (d) the class was not inferred but simply plotted from the known labels, providing a visual guide as to how VAE analysis without labels would perform.

We further extended this approach to experimental data, namely the analysis of the gold nanoparticle (GNPs) assemblies. The GNPs were salted with NaCl solution to an indium-tin oxide (ITO) substrate. The process of salting was observed using dark field microscopy to make sure enough GNPs are on the ITO substrate. Then, we performed imaging on the same area using scanning electron microscopy (SEM). The nanoparticle dataset is automatically created from the SEM micrographs and then hand-labeled manually by us using the number particles in each image as "class".

Here, we limited ourselves only to classes that have no less than 30 "samples" (images). This left us with four classes corresponding to 1-, 2-, 3- and 4-particle agglomerations, with 304, 93, 49, and 31 images in each class. We took 15 samples from each class to prepare the labeled dataset, whereas the remaining ones went into the unlabeled dataset. Note that this created a significant class imbalance against which we are going to test our ss-rVAE model.

As initial test, we attempted the simple VAE and rVAE approaches (i.e., z-Encoder and Decoder only, without (VAE) and with (rVAE) partitioning of the continuous latent z-space). Figure 5 shows the latent space and distribution of the encoded latent variables of VAE and rVAE analyses. In this case, the training is completely unsupervised and no classification is performed (albeit the distribution of the data in the latent space can provide indication on feasibility of the latter). One can see an evolution of particle numbers along the vertical direction in both Figure 5a and 5b. In addition, by encoding the entire dataset into the latent space, we can see a cluster(s) of points in the latent space corresponding to images with a single particle, indicating the successful classification of the latter (Figure 5c and 5d). However, for VAE analysis (Figure 5c), all other labels are mixed, suggesting a failure of further distinguishing other classes. In this regard, the rVAE, which has a rotational invariance, performs slightly better (Figure 5d) but this performance remains insufficient to classify other images. In addition, it is observed that the reconstruction performance of both the VAE and rVAE for the images with 3 and 4 GNPs is very poor, as seen in the bottom part of Figure 5a-b.

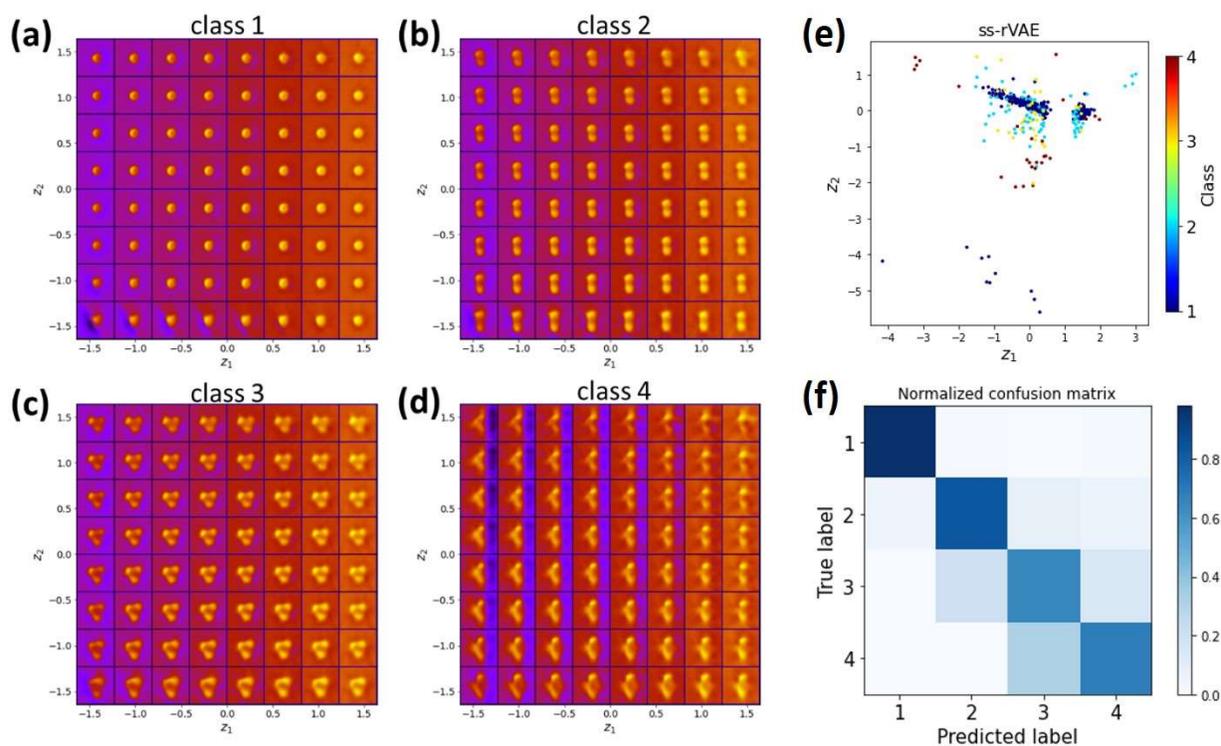

**Figure 6. ss-rVAE analysis of the nanoparticle dataset. (a-d)** class-conditioned latent space of the ss-rVAE trained on partially labeled GNP dataset. **(e)** latent variables distribution with color corresponding to the class variable. Note that the discrete class and continuous latent variables are encoded with two different encoders ($y$-Encoder and $z$-Encoder) and therefore the class variable forms a separate dimension, which in this case is collapsed onto the 2D plane of the continuous latent representation. (f) Confusion matrix for the predictions of the trained classifier on the unlabeled part of the dataset.

Hence, we applied the ss-rVAE to the GNPs dataset. In this case, the training is semi-supervised and the classification is performed via the *y*-Encoder by encoding the data into a discrete latent variable. Then, other information, such as particle size and background signal intensity, is encoded into the standard continuous latent variables that contain information on physical factors of variability in data. The class-conditioned continuous latent spaces shown in Figure 6a-d indicate good performance of ss-rVAE in the classification task as well as in discovering the main factors of variation for the images associated with each individual class. Indeed, both particle number and particle shapes are well identified in the latent space of the first three classes (Figure 6a-c). The blurring of the latent space for the fourth class is due to the large number of possible configurations and a (very) small number of examples. We also observe a variation of particle size and background information, suggesting that the physical meaning of the conventional latent variables is related to the particle size and background. The comparison of VAE, rVAE, and ss-rVAE clearly indicates the superior performance of the ss-rVAE.

To summarize, we have introduced a semi-supervised rotationally-invariant variational autoencoder (ss-rVAE) as a universal approach that allows generalizing from a small subset of labeled data with a weak orientational disorder to a large unlabeled dataset with a much stronger orientational disorder. This approach both allows recovering missing labels for a dataset and disentangling continuous factors of variation for each class. Finally, the classifier part of the trained ss-rVAE inherits the rotational (and translational) invariances and can be deployed independently of the other parts of the model. The performance of the ss-rVAE was illustrated using synthetic data sets with the known factors of variation and was further extended to experimental data sets of clusters of gold nanoparticles. Hence, this approach provides a universal framework for the analysis of imaging data in areas as diversified as biology, medicine, condensed matter physics and materials science, and techniques ranging from optical to electron and scanning probe microscopies. In particular, it directly maps the common situation when small amounts of labeled data are available, in the form of prior published results, image collections and catalogs, or even theoretical models. It also can significantly reduce the need for labeled data. Finally, we note that VAE approach is naturally extendable for defining complex physical phenomena such as causal relationships (in the form of directed acyclic graphs connecting latent variables), topological structure of the data (via invariances of the latent space), opening the pathway towards exploring these phenomena based on observational data.

**Acknowledgment:**

This work was supported (M.Y.Y., D.G., S.V.K.) by the US Department of Energy, Office of Science, Office of Basic Energy Sciences, as part of the Energy Frontier Research Centers program: CSSAS–The Center for the Science of Synthesis Across Scales–under Award Number DE-SC0019288, located at University of Washington and performed (M.Z.) at Oak Ridge National Laboratory's Center for Nanophase Materials Sciences (CNMS), a U.S. Department of Energy, Office of Science User Facility. SEM imaging was conducted at the University of Washington

Molecular Analysis Facility, a National Nanotechnology Coordinated Infrastructure (NNCI) site which is supported in part by the National Science Foundation, the University of Washington, the Molecular Engineering and Sciences Institute, and the Clean Energy Institute. D.S.G. acknowledges support from the University of Washington, Department of Chemistry Kwiram Endowment. The ML of experimental data is supported (Y.L.) by the U.S. Department of Energy, Office of Science, Office of Basic Energy Sciences Energy Frontier Research Centers program under Award Number DE-SC0021118. The authors acknowledge Dr. Ilia Ivanov (CNMS) for early ideas (~2010) of creating nanoparticle libraries.

**Code and data availability:**

Code and data are available without restrictions from https://github.com/ziatdinovmax/Semi-Supervised-VAE-nanoparticles